\pdfoutput=1

\documentclass[11pt]{article}

\usepackage[]{acl}

\usepackage{times}
\usepackage{soul}
\usepackage{url}
\usepackage{amsmath}
\usepackage{amsthm}
\usepackage{booktabs}
\usepackage{algorithm}
\usepackage{algorithmic}
\usepackage[switch]{lineno}

\usepackage{latexsym}
\usepackage{braket}
\usepackage{amssymb}
\usepackage{multirow}
\usepackage{subcaption}
\usepackage[normalem]{ulem}
\useunder{\uline}{\ul}{}

\usepackage[T1]{fontenc}

\usepackage[utf8]{inputenc}

\usepackage{microtype}

\usepackage{inconsolata}

\usepackage{graphicx}

\usepackage{setspace}
\title{A Systematic Analysis on the Temporal Generalization \\ of Language Models in Social Media}

\author{
    Asahi Ushio \and Jose Camacho-Collados \\
    Cardiff NLP, Cardiff University, UK \\
    \tt \{ushioa,camachocolladosj\}@cardiff.ac.uk \\
}

\begin{document}
\maketitle

\begin{spacing}{0.98}

\begin{abstract}
In machine learning, temporal shifts occur when there are differences between training and test splits in terms of time. For streaming data such as news or social media, models are commonly trained on a fixed corpus from a certain period of time, and they can become obsolete due to the dynamism and evolving nature of online content. This paper focuses on temporal shifts in social media and, in particular, Twitter. We propose a unified evaluation scheme to assess the performance of language models (LMs) under temporal shift on standard social media tasks. LMs are tested on five diverse social media NLP tasks under different temporal settings, which revealed two important findings: (i) the decrease in performance under temporal shift is consistent across different models for entity-focused tasks such as named entity recognition or disambiguation, and hate speech detection, but not significant in the other tasks analysed (i.e., topic and sentiment classification); and (ii) continuous pre-training on the test period does not improve the temporal adaptability of LMs.
\end{abstract}

\section{Introduction}
Modern natural language processing (NLP) is centered on language models (LMs)~\cite{devlin-etal-2019-bert-short,GPT2,liu2019roberta,min2023recent}. The versatility of LMs has enabled many real world applications, including chatbot\footnote{\url{https://openai.com/blog/chatgpt}}, text-guided image generation~\cite{ramesh2021zeroshot}, and text-to-speech~\cite{AudioPaLM}. One of the well-known issues of LMs, however, is that the capabilities of LMs can not be fully analyzed due to their blackbox nature. To overcome such limitations to understand LMs' true capability, methodologies and datasets to inspect LMs have been proposed in the context of model probing study, which uncovered various features such as syntax~\cite{hewitt-manning-2019-structural,goldberg2019assessing}, factual knowledge~\cite{petroni-etal-2019-language-short,ushio-etal-2021-bert-short}, semantics~\cite{ettinger-2020-bert,tenney-etal-2019-bert}, and emergent ability~\cite{wei2022emergent}.

Besides such studies of LM probing, there is another line of research that focuses on the adaptability of LMs under settings incurring changing conditions, including \emph{temporal shifts}~\cite{NEURIPS2021_f5bf0ba0,loureiro-etal-2022-timelms-short}. In this paper, we refer to temporal shifts when discussing settings in which the time period of the test set is different from that of the training set (with the test set period being generally \textit{after}, reassembling real-world settings.). These settings have been empirically known to lead a non-trivial decrease in performance on some tasks~\cite{liska2022streamingqa,kasai2022realtime}. Needless to say, temporal shifts are more important in more dynamic streaming data with frequent meaning changes and evolving entities, such as social media~\cite{antypas-etal-2022-twitter-short,ushio-etal-2022-named-short}.

In this paper, we focus on temporal shifts on Twitter, one of the major social media platforms, and propose a unified evaluation scheme to assess the adaptability of LMs toward temporal shift on Twitter. In particular, we are interested in answering the following two research questions: 
\begin{itemize}
    \item \textbf{RQ1.} Are temporal shifts in social media detrimental for LM performance in NLP tasks?
    \item \textbf{RQ2.} If so, what are the causes of this temporal shift and can it be mitigated (by e.g. using LMs pre-trained on recent data)?
\end{itemize}

For the evaluation, we selected five diverse social media NLP tasks for which there are datasets with temporal information available: hate speech detection, topic classification, sentiment classification, named entity disambiguation (NED), and named entity recognition (NER) ranging over different time periods. The temporal shifts considered are relatively short compared to those studied in other sources of streaming data such as news and scientific papers. We test both LMs specialized on social media and other general-purpose trained on encyclopedic and web-crawled corpus.

Our study shows that tasks driven by named entities or events (i.e., hate speech, NED, and NER) present consistent decrease across model under temporal shift settings, while it is less prominent in the other tasks. Crucially, our results show that the decrease caused by temporal shift cannot be mitigated by considering a more recent corpus to the pre-training dataset. Finally, qualitative analysis highlights that the common mistakes made by LMs are indeed instances that require to understand the named entities in the tweet. 
All the datasets and the scripts to reproduce our experiments are made publicly available online\footnote{\url{https://huggingface.co/datasets/tweettemposhift/tweet_temporal_shift}}.

\section{Related Work}
\label{sec:related-work}
\paragraph{LMs on Social Media.} Major LMs are commonly pre-trained on encyclopedic and web-crawled corpora~\cite{lewis-etal-2020-bart-short,T5,PaLM,PaLM2,Llama,Llama2,GPT3}, while the adaptation of such LMs to social media has led new LMs pre-trained on corpus curated over social media~\cite{nguyen-etal-2020-bertweet,loureiro-etal-2022-timelms-short,delucia-etal-2022-bernice,barbieri-etal-2022-xlm-short}, which present better performance on social media NLP tasks than standard LMs~\cite{barbieri-etal-2020-tweeteval,antypas-etal-2023-supertweeteval}. However, such studies on NLP tasks in social media mainly focus on static datasets without temporal shift. A few of them associate timestamps to the dataset and provide basic temporal analysis~\cite{antypas-etal-2022-twitter-short,ushio-etal-2022-named-short}, but these are limited to a single task. Finally, related to the temporal aspect of this work, short-term meaning shift has also been studied in the context of social media and LMs \cite{loureiro-etal-2022-tempowic-short}.

\paragraph{Temporal Generalization.} Importantly, this work aligns to the research on the temporal or diachronic generalization of LMs. In this area, however, most previous works focus on relatively long term (over 10 years)~\cite{NEURIPS2021_f5bf0ba0} or formal source of text such as news and scientific papers~\cite{liska2022streamingqa,kasai2022realtime}. In the context of short-term temporal analysis, there are three studies that are most similar to ours. \citeauthor{luu-etal-2022-time-custom} (\citeyear{luu-etal-2022-time-custom}) analyse the temporal performance degradation of LMs in NLP tasks in relatively short time periods. While social media is included as one of the domains, the evaluation is limited to the classification task and to general-domain models. \citeauthor{agarwal-nenkova-2022-temporal} (\citeyear{agarwal-nenkova-2022-temporal}) performed a similar general analysis for different tasks, while also analysing the effect of self-labeling as a mitigation to temporal misalignment, which we do not analyse in this work. The main difference between these works in ours is our focus on social media, where we carry out a targeted comprehensive analysis on short-term temporal effects. When it comes to social media, temporal shifts are especially relevant given the real-time nature of the platform and their focus on current events. In the context of Italian Twitter, \citeauthor{florio2020time} (\citeyear{florio2020time}) analysed the temporal sensitivity of models for hate speech detection, which is one of the tasks included in this paper.

\paragraph{Temporal-aware LMs.} To enhance adaptability of LMs for temporal shift, there are a few works that explicitly ingests the temporal information to the model by specific attention mechanism~\cite{rosin-radinsky-2022-temporal}, augmenting the input with timestamp~\cite{10.1145/3488560.3498529}, joint modeling of temporal information~\cite{dhingra-etal-2022-time}, and self-labeling~\cite{agarwal-nenkova-2022-temporal}. In this paper, we do not include any temporal-aware LMs, because we are interested in analysing the adaptability of plain LMs to temporal shifts.


\section{Experimental Setting}

In this section, we describe our experimental setting to investigate the effect of temporal shifts in LMs.

\subsection{Evaluation Methodology}

Let us define $\mathcal{D}_{{\rm train}}$ and
$\mathcal{D}_{{\rm test}}$ as the training and test splits of a dataset $\mathcal{D}$ for a single downstream task (e.g. sentiment classification), where each dataset contains pairs of a text input and associated labels. Importantly, $\mathcal{D}_{{\rm train}}$ is taken from the period prior to $\mathcal{D}_{{\rm test}}$ without any temporal overlap. Given such dataset with temporal split, we consider the following two settings of out-of-time (OOT) and in-time (IT).

\paragraph{Out-of-Time (OOT).}
In the first setting, we simply train the models on $\mathcal{D}_{{\rm train}}$ and evaluate them on $\mathcal{D}_{{\rm test}}$. Noticeably, models have no access to the instances from the test period at the training phase in this setting, so we refer the setting as \emph{out-of-time} (OOT) as an analogy to the out-of-domain (OOD).

\begin{figure}[!t]
\centering
\includegraphics[width=0.8\columnwidth]{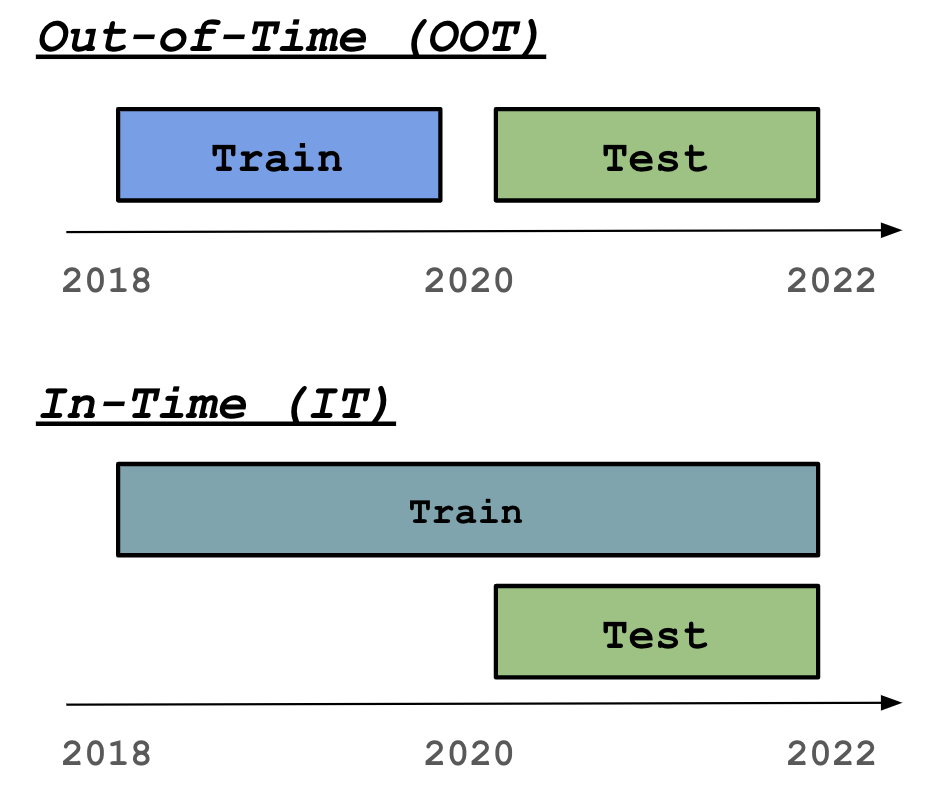}
\caption{An illustrative example of the conceptual differences between the sampling time periods of the OOT and IT settings.}
\label{fig:oot}
\end{figure}

\begin{table*}[!t]
    \centering
    \scalebox{0.8}{
    \begin{tabular}{lrrrp{300pt}}
    \toprule
       &  Split &    Size &                     Date  & Examples\\
    \midrule
    \multirow{3}{*}{\rotatebox{90}{Hate}}       
&  Train &   2,318 &  2013-09-23 / 2015-03-03 & \emph{Zebra undies \#MKR chic in pink dress.} (Hate)\\
&  Valid &     579 &  2013-09-23 / 2015-03-03 & \emph{OMG fashion parade time \#mkr.} (non-Hate)\\
&   Test &   1,475 &  2015-03-04 / 2015-03-14 &  \emph{female football commentators just don't work.} (Hate) \\
          \midrule
        \multirow{3}{*}{\rotatebox{90}{Topic}}     
&  Train &   4,585 &  2019-09-08 / 2020-08-30 & \emph{So, when I can listen to watermelon sugar live in Jakarta Harry?}\\
&  Valid &     573 &  2019-09-08 / 2020-08-30 & \emph{@Harry\_Styles} (celebrity, music)	\\
&   Test &   1,679 &  2020-09-06 / 2021-08-29 & \emph{Glad to see the Chiefs crushed the Texans} (sports)\\ \midrule
    \multirow{3}{*}{\rotatebox{90}{Sent.}} 
&  Train &   5,000 &  2014-02-06 / 2016-12-31 & \emph{I think I'm in love} (positive)\\
&  Valid &   1,344 &  2016-01-01 / 2016-12-31 & \emph{@user is making me very upset} (negative)\\
&   Test &   1,344 &  2018-01-01 / 2019-01-01 & \emph{Shoutout to @MENTION for donating to poor} (positive) \\\midrule
    \multirow{4}{*}{\rotatebox{90}{NED}}       
&  Train &  18,469 &  2020-02-26 / 2021-08-27 & \emph{Every concert I’ve seen announce lately, they are steering clear of Detroit} (Target: Detroit, Definition: Art museum, Label: False) \\
&  Valid &   4,617 &  2020-02-27 / 2021-08-27 & \emph{Me on stream: Happy Friday!, Australia: It’s Saturday} \\
&   Test &  21,253 &  2021-08-28 / 2021-11-28 & (Target: Australia, Definition: country, Label: True)\\ \midrule
    \multirow{3}{*}{\rotatebox{90}{NER}}       
&  Train &   4,616 &  2019-09-08 / 2020-08-30 & \emph{UFC 245: Looking at the three title fights on tap at T-Mobile Arena} \\
&  Valid &     576 &  2019-09-08 / 2020-08-30 & (UFC 245: corporation, T-Mobile Arena: location)\\
&   Test &   2,807 &  2020-09-05 / 2021-08-31 & \emph{Glad the Chiefs crushed the Texans} (Chiefs: group, Texans: group) \\
    \bottomrule
    \end{tabular}
    }
    \caption{The number of tweets and the period with examples of each dataset.}
    \label{tab:stats}
\end{table*}

\paragraph{In-Time (IT).}
As a comparison to OOT, we consider the second experimental setting, which is designed to assess the effect of training instances from the test period. 
The test set is randomly split into non-overlapped four chunks ($\mathcal{D}_{{\rm test}}=\bigcup_{i=1}^4\mathcal{D}_{{\rm test}}^i$) for cross validation, where models trained on $\mathcal{D}_{{\rm train}} \bigcup \mathcal{D}_{{\rm test}}\backslash \mathcal{D}_{{\rm test}}^i$ are evaluated on $\mathcal{D}_{{\rm test}}^i$.
For each chunk of the test set, we downsample the IT training set to the same size as $\mathcal{D}_{{\rm train}}$ with three random seeds and report the averaged metrics over the runs. To be precise, we consider a function $\mathcal{F}_s(\mathcal{D})$ that randomly samples $|\mathcal{D}_{\rm train}|$ instances from $\mathcal{D}$, and we independently train models on $\mathcal{F}_s(\mathcal{D}_{{\rm train}} \bigcup \mathcal{D}_{{\rm test}}\backslash \mathcal{D}_{{\rm test}}^i)$ for 
$s=0,1,2$. In contrast to OOT, we refer this setting as \emph{in-time} (IT) setting.

Figure~\ref{fig:oot} presents an example overview of the differences between IT and OOT settings from the perspective of data sampling periods (data from 2018 to 2022 in the example).

\subsection{Tasks \& Datasets}
We consider the following five diverse social media NLP tasks: hate speech detection, topic classification, sentiment classification, named entity disambiguation (NED), and named entity recognition (NER). For each task, we employ a public dataset for English and leverage its original temporal splits, unless there is overlap between the periods of training and test sets.

\paragraph{Hate Speech Detection.}
Hate speech detection in Twitter consists of identifying whether a tweet contains hateful content. We use the dataset proposed by Waseem and Hovy~\shortcite{waseem-hovy-2016-hateful} framed as binary classification as the dataset to create the training and test splits based on the timestamp. The first half is used for training and the rest for test split. The training split is further randomly split into 2:8 for validation:training. We use accuracy to evaluate the hate speech detection models.

\paragraph{Topic Classification.}
Topic classification is a task that consists of associating an input text with one or more labels from a fixed label set. For this evaluation, we rely on TweetTopic~\cite{antypas-etal-2022-twitter-short}, a multi-label topic classification dataset with 19 topics such as \textit{sports} or \textit{music}. As evaluation metric, we use micro-F1 score to measure the performance of topic classification models.

\paragraph{Sentiment Analysis.}
Sentiment analysis is a standard social media task consisting of associating each post with its sentiment. In particular, we use the dataset from the task 2: LongEval-Classification from CLEF-2023~\cite{alkhalifa2023overview} in which the task is framed as binary classification with positive and negative labels. The original training split contains around 50k instances while 1k test split, which is highly imbalance and the effect of the IT sample can be very limited. To balance the training and test splits, we randomly sample 2.5k instances from each label, amounting 5k new training instance. We use accuracy to evaluate the sentiment classification models.

\paragraph{Named Entity Disambiguation (NED).}
NED is a a binary classification that consists of identifying if the meaning of a given target entity in context is the same as the one provided. We use the TweetNERD~\cite{mishra2022tweetnerd} 
dataset and reformulated into NED following SuperTweetEval ~\cite{antypas2023supertweeteval}. Then, we create the train, validation, and test splits in the same way as the hate speech detection. We use accuracy to evaluate the NED models.

\paragraph{Named Entity Recognition (NER).}
NER is a sequence labelling task to predict a single named-entity type on each token on the input text. We rely on TweetNER7~\cite{ushio-etal-2022-named-short}, a NER dataset on Twitter that contains seven named entity types. We use span F1 score to evaluate NER models.

\subsubsection{Data Statistics}

\begin{figure}[!t]
\centering
\includegraphics[width=\columnwidth]{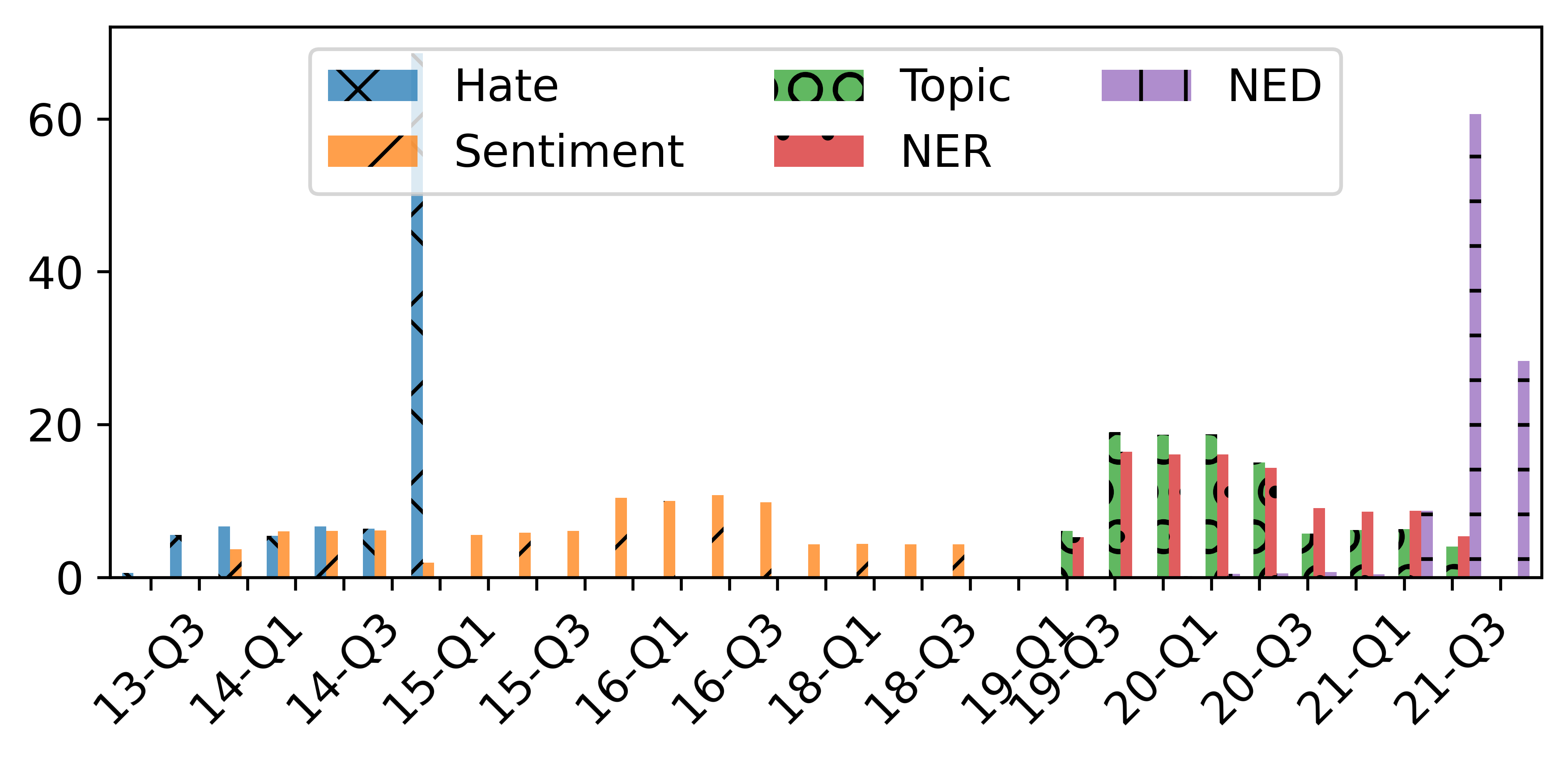}
\caption{Quarterly breakdown of the number of tweets ratio (\%) in each dataset. For example, a ratio of 5\% in 13-Q3 for Dataset X would mean that 5\% of all tweets in Dataset X belong to the third quarter (July-September) of 2013.}
\label{fig:data_period}
\end{figure}

\begin{figure}[!t]
    \centering
    \includegraphics[width=\columnwidth]{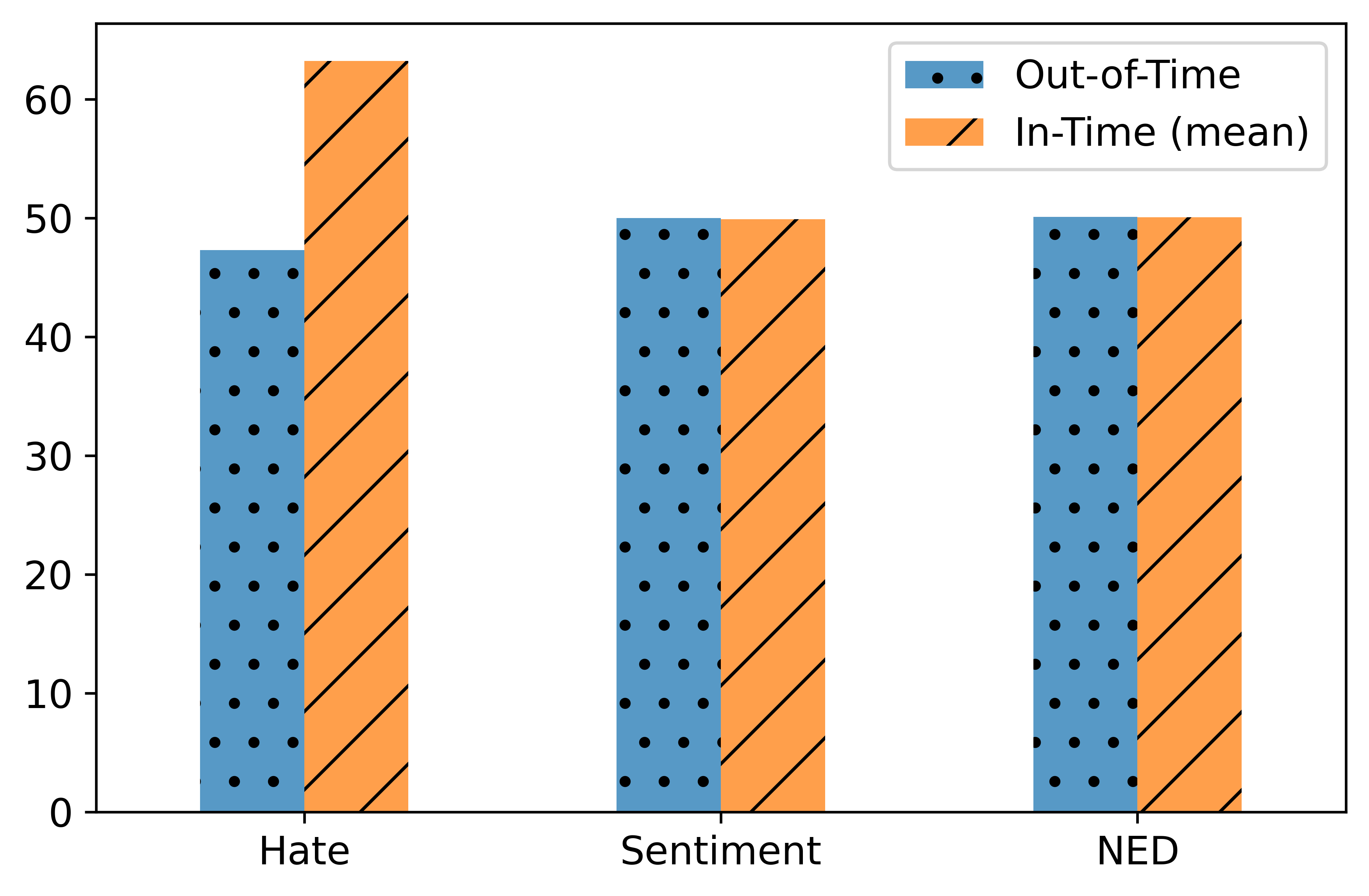}
    \caption{Comparisons of ratio (\%) of positive labels in the training split of each task between OOT and IT.}
    \label{fig:label-dist-binary}
\end{figure}

\begin{figure}[t]
     \centering
     \begin{subfigure}[b]{\columnwidth}
         \centering
         \includegraphics[width=\textwidth]{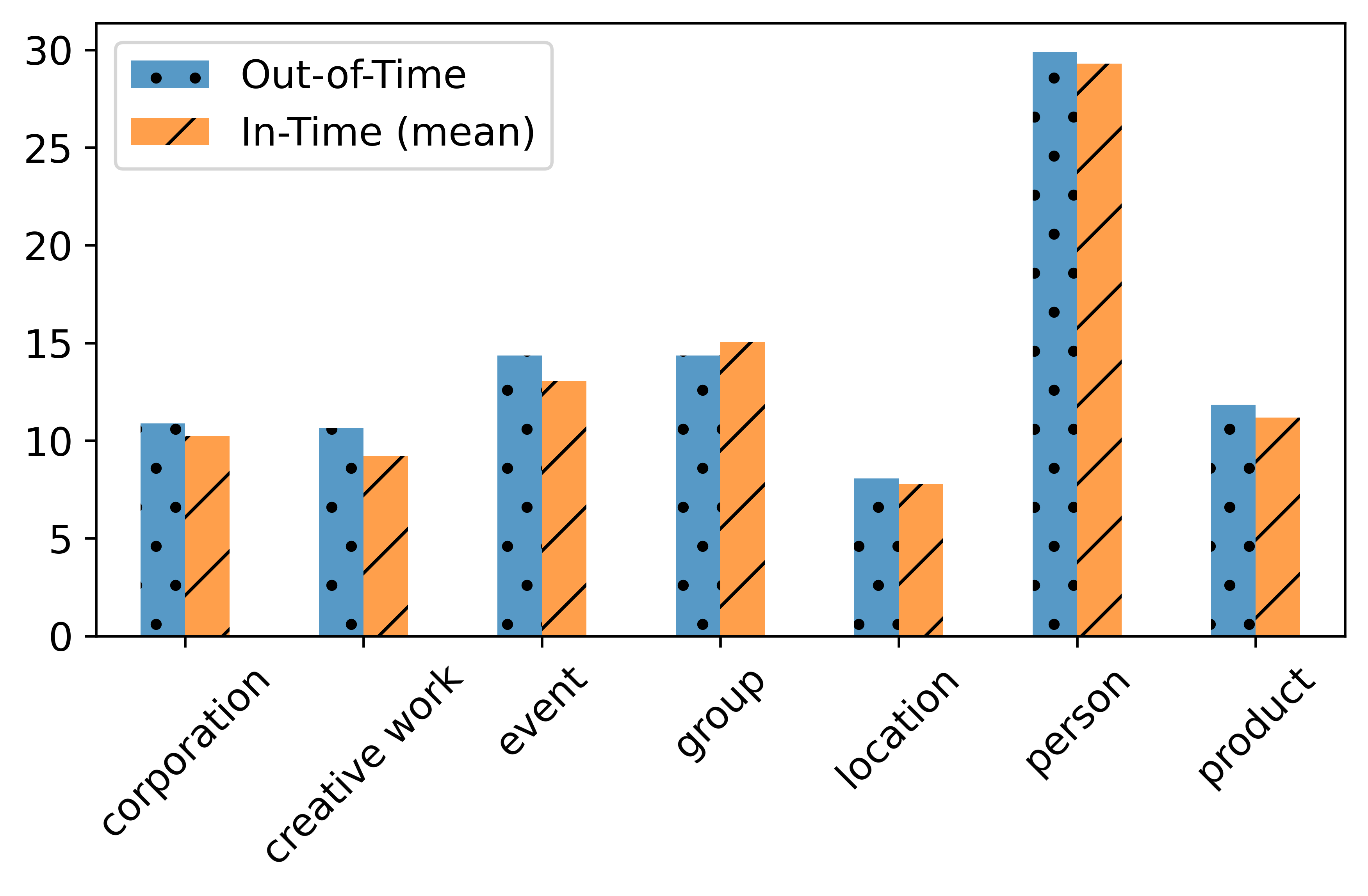}
         \caption{Ratio of entities in NER.}
     \end{subfigure}
     \hfill
     \begin{subfigure}[b]{\columnwidth}
         \centering
         \includegraphics[width=\textwidth]{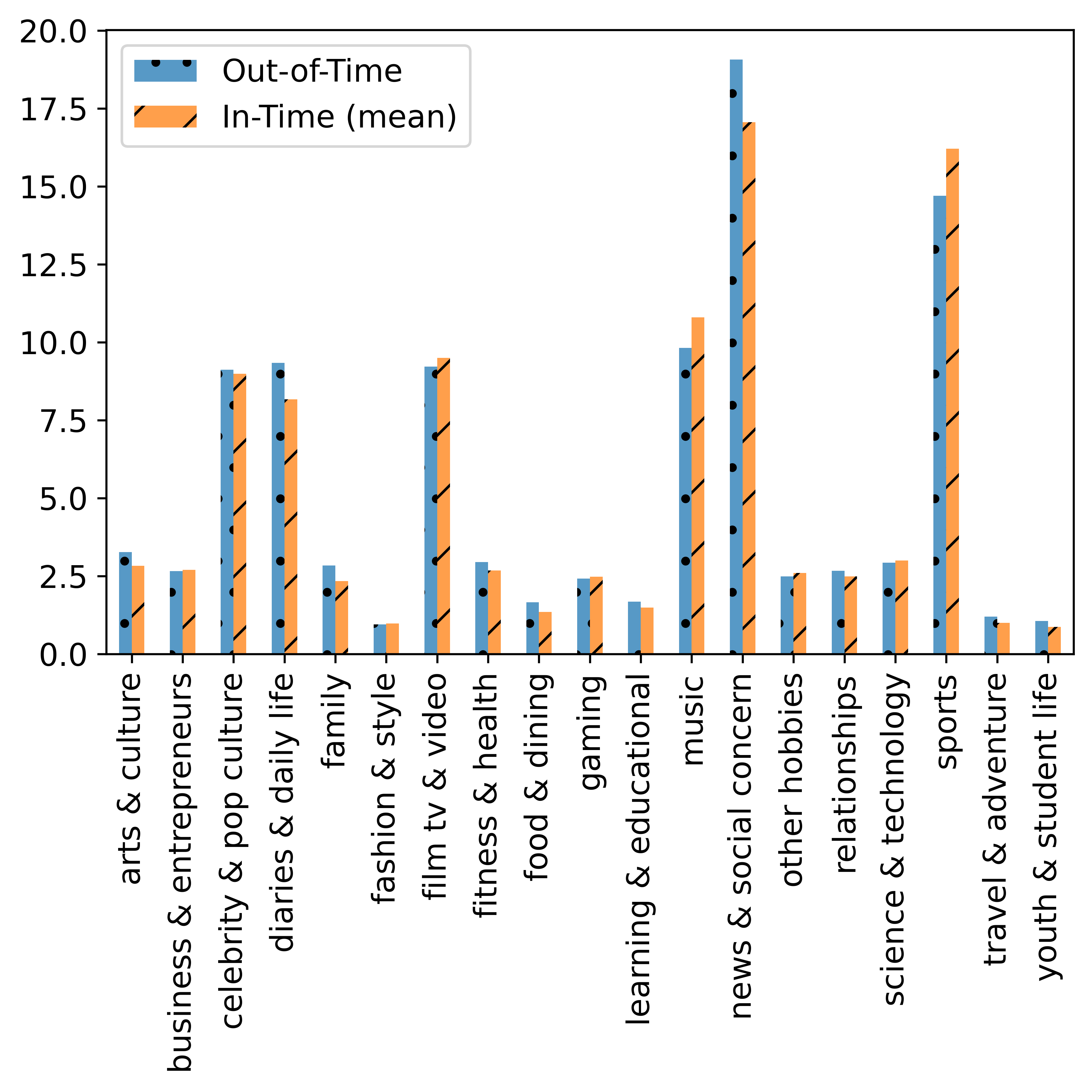}
         \caption{Ratio of labels in topic classification.}
     \end{subfigure}
\caption{Comparisons of label distributions between OOT and IT settings.}
\label{fig:label-dist}
\end{figure}

Table~\ref{tab:stats} shows the size and the period of the training and the test sets for each dataset, and Figure~\ref{fig:data_period} displays the number of tweets per quarter for each task.
Topic classification and NER use the same tweets, which are sampled uniformly from each month, while NED and hate speech detection have the majority of the tweets in the latest quarter. Sentiment analysis covers the longest period in the dataset that spans over four years.
Figures~\ref{fig:label-dist-binary} and \ref{fig:label-dist} show the comparisons of the label distribution of the binary (i.e., hate speech, sentiment classification, and NED) and multi-classification tasks (i.e., NER and topic classification), respectively. 
As can be observed, hate speech detection has fewer positive labels in OOT than in IT, while the other two tasks have the same ratio of the positive labels between OOT and IT. The same pattern can be observed for topic classification and NED, for which the label distribution does not substantially change.

\begin{table*}[t]
\scalebox{0.8}{
\begin{tabular}{llll}
\toprule
Model                           & Parameters & HF Name & Citation \\ \midrule
RoBERTa\textsubscript{BASE}     & 123M & \texttt{roberta-base}                              & \multirow{2}{*}{\cite{liu2019roberta}} \\ \cmidrule{1-3}
RoBERTa\textsubscript{LARGE}    & 354M & \texttt{roberta-large}                             & \\ \midrule
BERTweet\textsubscript{BASE}    & 123M & \texttt{vinai/bertweet-base}                              & \multirow{2}{*}{\cite{nguyen-etal-2020-bertweet}} \\ \cmidrule{1-3}
BERTweet\textsubscript{LARGE}   & 354M & \texttt{vinai/bertweet-large}                             & \\ \midrule
TimeLM2019\textsubscript{BASE}  & 123M & \texttt{cardiffnlp/twitter-roberta-base-2019-90m}  & \multirow{6}{*}{\cite{loureiro-etal-2022-timelms-short}} \\ \cmidrule{1-3}
TimeLM2020\textsubscript{BASE}  & 123M & \texttt{cardiffnlp/twitter-roberta-base-dec2020}   & \\ \cmidrule{1-3}
TimeLM2021\textsubscript{BASE}  & 123M & \texttt{cardiffnlp/twitter-roberta-base-2021-124m} & \\ \cmidrule{1-3}
TimeLM2022\textsubscript{BASE}  & 354M & \texttt{cardiffnlp/twitter-roberta-base-2022-154m} & \\ \cmidrule{1-3}
TimeLM2022\textsubscript{LARGE} & 354M & \texttt{cardiffnlp/twitter-roberta-large-2022-154m}& \\ \midrule
BERNICE                         & 278M & \texttt{jhu-clsp/bernice}                          &\cite{delucia-etal-2022-bernice} \\
\bottomrule
\end{tabular}
}
\caption{Language models used in the paper with the number of parameters and model aliases on Hugging Face.}
\label{tab:model_details}
\end{table*}

\subsection{Models}

\begin{table}[t]
    \centering
    \scalebox{0.9}{
\begin{tabular}{@{}l@{\hspace{10pt}}c@{\hspace{10pt}}c@{\hspace{10pt}}c@{\hspace{10pt}}c@{\hspace{10pt}}c@{}}
\toprule
           & Hate     & Topic    & Sentiment    & NED      & NER\\ \midrule
RoBERTa    &\checkmark&          &\checkmark&          & \\
BERTweet   &\checkmark&          &\checkmark&          & \\
BERNICE    &\checkmark&\checkmark&\checkmark&\checkmark&\checkmark \\
TimeLM2019 &\checkmark&          &\checkmark&          & \\
TimeLM2020 &\checkmark&          &\checkmark&          & \\
TimeLM2021 &\checkmark&\checkmark&\checkmark&\checkmark&\checkmark \\
TimeLM2022 &\checkmark&\checkmark&\checkmark&\checkmark&\checkmark \\ \bottomrule
\end{tabular}
    }
\caption{The overlap between the test period and the pre-trained corpus of each LM (\checkmark indicates that the LM is pre-trained on the corpus including the test period of the task).}
\label{tab:data-period}
\end{table}

We investigate an established general-purpose LM, RoBERTa~\cite{liu2019roberta} as well as other LMs pre-trained on tweets including BERTweet~\cite{nguyen-etal-2020-bertweet}, TimeLM~\cite{loureiro-etal-2022-timelms-short}, and BERNICE~\cite{delucia-etal-2022-bernice}. For RoBERTa and BERTweet, we consider the base and the large models, referred as RoBERTa (B), RoBERTa (L), BERTweet (B), and BERTweet (L). For TimeLM, we consider the base models trained on the tweets up to 2019, 2020, 2021 and 2022, referred as TimeLM2019 (B), TimeLM2020 (B), TimeLM2021 (B) and TimeLM2022 (B), and the large model trained upto 2022, referred as TimeLM2022 (L). The end date of the pre-trained corpus for each model is 2019-02 (RoBERTa), 2019-08 (BERTweet), 2019-12 (TimeLM2019), 2020-12 (TimeLM2020), 2021-12 (TimeLM2021 and BERNICE), and 2022-12 (TimeLM2022).
All the model weights are taken from the transformers model hub~\cite{wolf-etal-2020-transformers-short} and Table~\ref{tab:model_details} shows the details of models we used in the paper.\footnote{Note that for this analysis we are not interested in the performance of zero-shot LLMs such as GPT-4, but rather on the effect of fine-tuned LMs.}
Table~\ref{tab:data-period} shows the overlap between the period of the pre-trained corpus and the test set for each task, which will be relevant for the analysis on the effect of pre-training in Section \ref{sec:pretraining}.
These models are then fine-tuned in the datasets presented in the previous section, in both OOT and IT settings. For model fine tuning, we run hyperparameter search with Optuna~\cite{akiba2019optuna} with the default search space. 

\section{Results}
\label{sec:result}

\begin{figure}[t]
\centering
\includegraphics[width=\columnwidth]{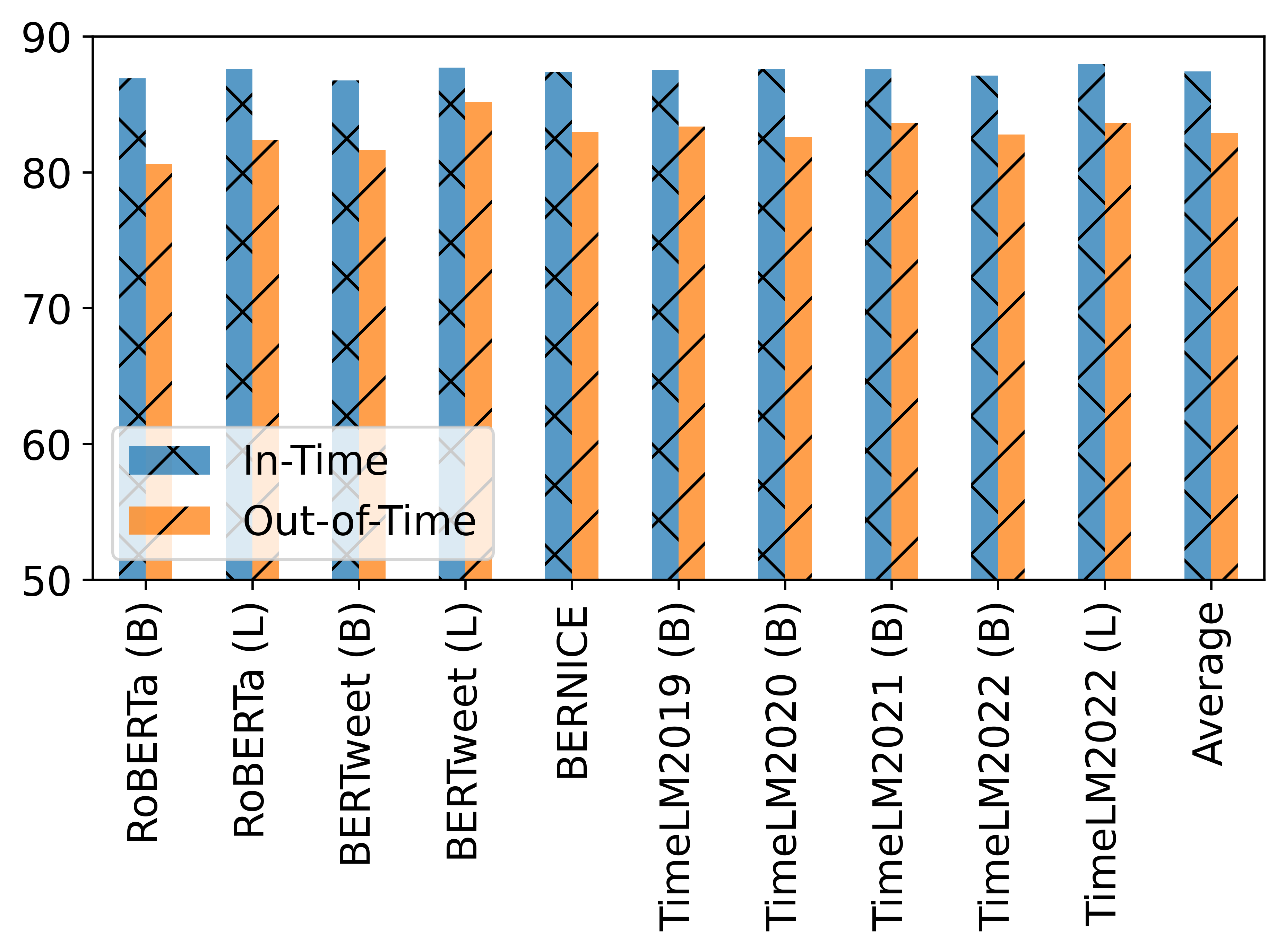}
\caption{Comparisons of IT and OOT performance (accuracy) for hate speech detection.}
\label{fig:result-bar-hate}
\end{figure}

\begin{figure}[t]
\centering
\includegraphics[width=\columnwidth]{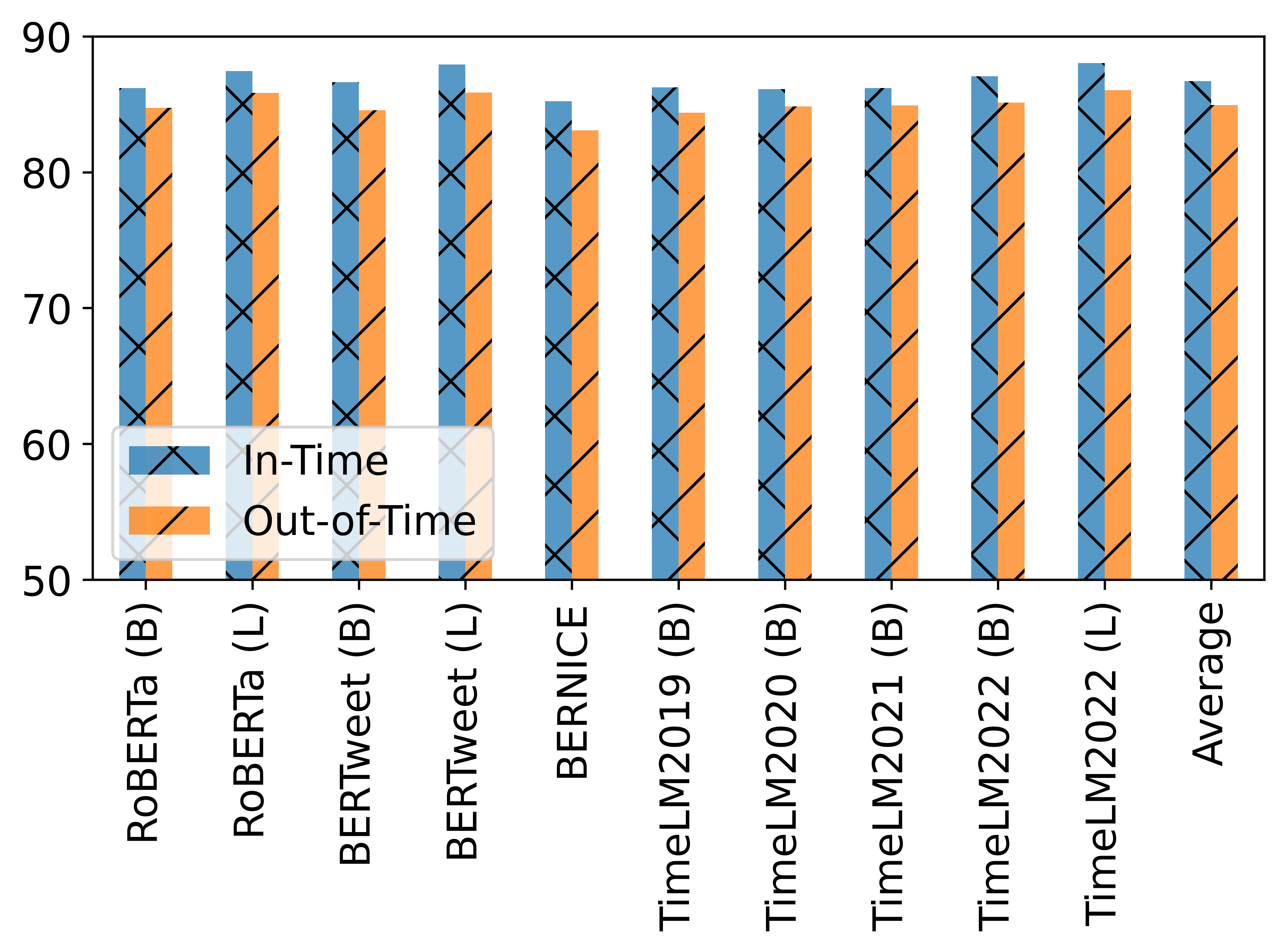}
\caption{Comparisons of IT and OOT performance (accuracy) for NED.}
\label{fig:result-bar-ned}
\end{figure}

\begin{figure}[t]
\centering
\includegraphics[width=\columnwidth]{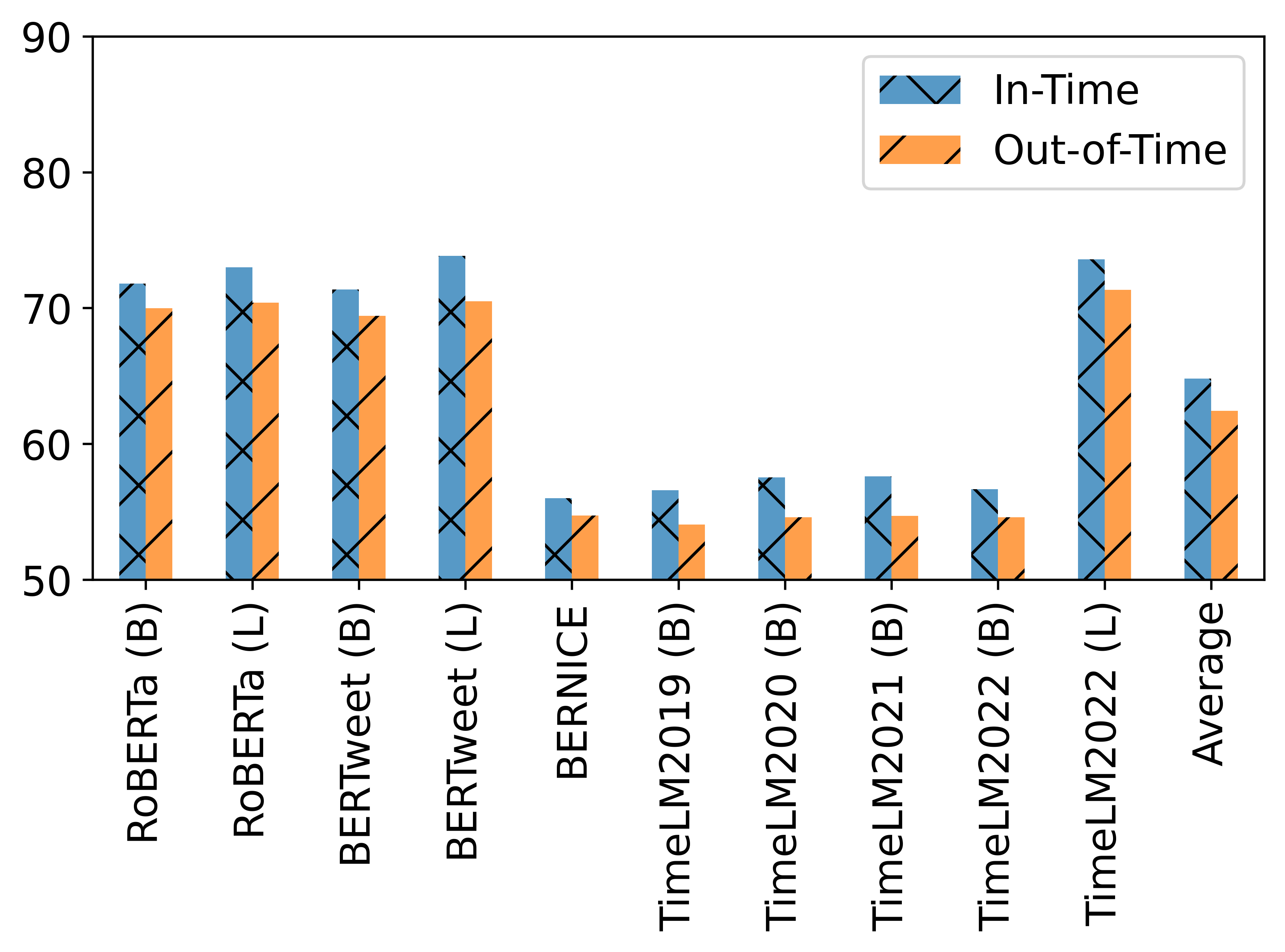}
\caption{Comparisons of IT and OOT performance (F1 score) for NER.}
\label{fig:result-bar-ner}
\end{figure}

\begin{figure}[t]
\centering
\includegraphics[width=\columnwidth]{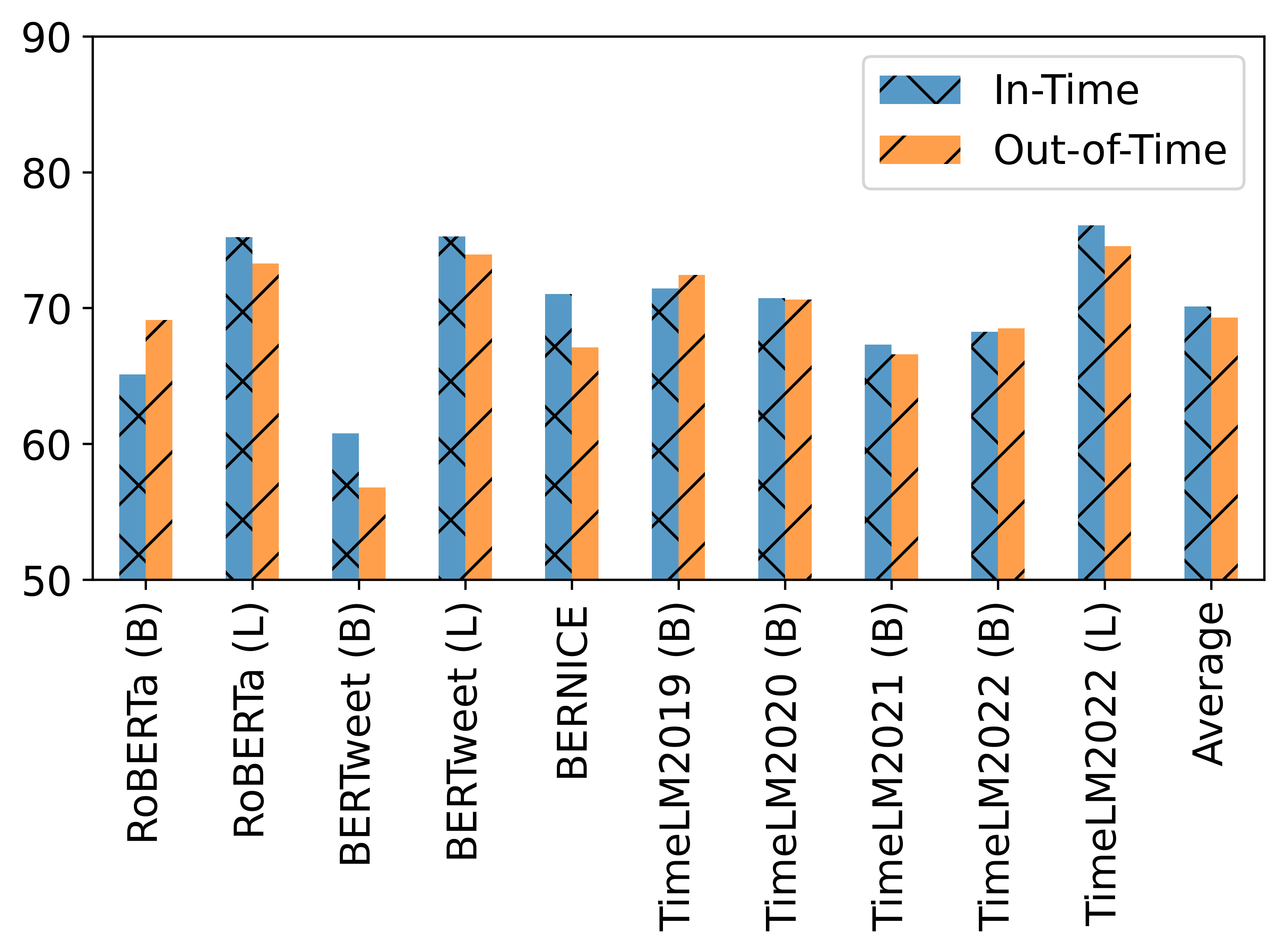}
\caption{Comparisons of IT and OOT performance (F1 score) for topic classification.}
\label{fig:result-bar-topic}
\end{figure}

\begin{figure}[t]
\centering
\includegraphics[width=\columnwidth]{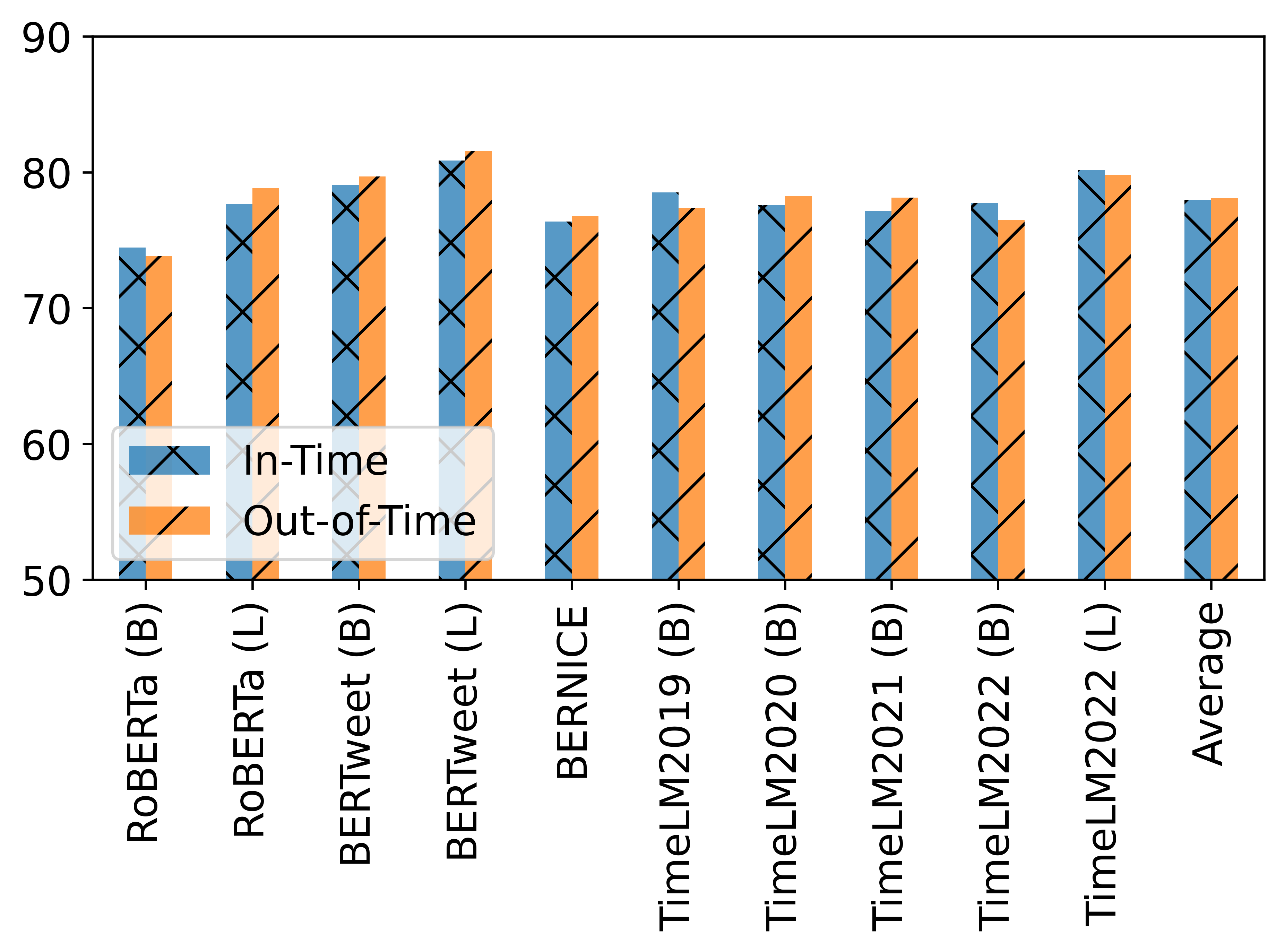}
\caption{Comparisons of IT and OOT performance (accuracy) for sentiment classification.}
\label{fig:result-bar-sentiment}
\end{figure}

Figures~\ref{fig:result-bar-hate} to \ref{fig:result-bar-sentiment} show the comparisons of IT and OOT in hate speech detection, NED, NER, topic classification and sentiment analysis. As can be observed, hate speech detection, NED and NER present inconsistencies in both settings, decreasing the performance from IT to OOT.
In contrast, this cannot be observed for both sentiment analysis, and especially topic classification. The average decrease of OOT performance for each of the tasks is 4.5, 2.4, 1.7, 0.8 and -0.1 for hate speech detection, NER, NED  topic classification and sentiment analysis.

One of the main differences of those two groups of tasks (i.e. hate/NED/NER v.s. topic/sentiment) entity-centric or event-driven nature of the former. NER and NED are clearly related to named entities. Hate speech detection does not relate to named entities explicitly, but since the tweets for hate speech detection are collected by querying specific events, they are often about events or celebrities which peak around the sampled timestamp \cite{gomez2023interaction}. 
On the other hand, events or named entities are not as important in sentiment analysis, as the sentiment can be estimated from the context in most cases. Topic classification depends on the topic, with some of them related to entities (e.g. those related to celebrities or TV) and others not (e.g., daily life, family or food), but in the main clearly identifiable by the context. Through the lens of entity relevancy, this result may suggest that the temporal shift can be caused by named entities, which includes meaning drift of existing named entities or emerging new named entities. Topic classification can be seen as a mixture of entity-related instances and not, which results in not fully consistent gain from OOT, but still significant in the average.

\section{Analysis}

This section focuses on the second research question (RQ2) and analyses the main causes behind temporal shift performance degradation of LMs.

\subsection{Effect of Pre-Training}
\label{sec:pretraining}

\begin{figure}[!t]
     \centering
     \hfill
     \begin{subfigure}[b]{\columnwidth}
         \centering
         \includegraphics[width=\textwidth]{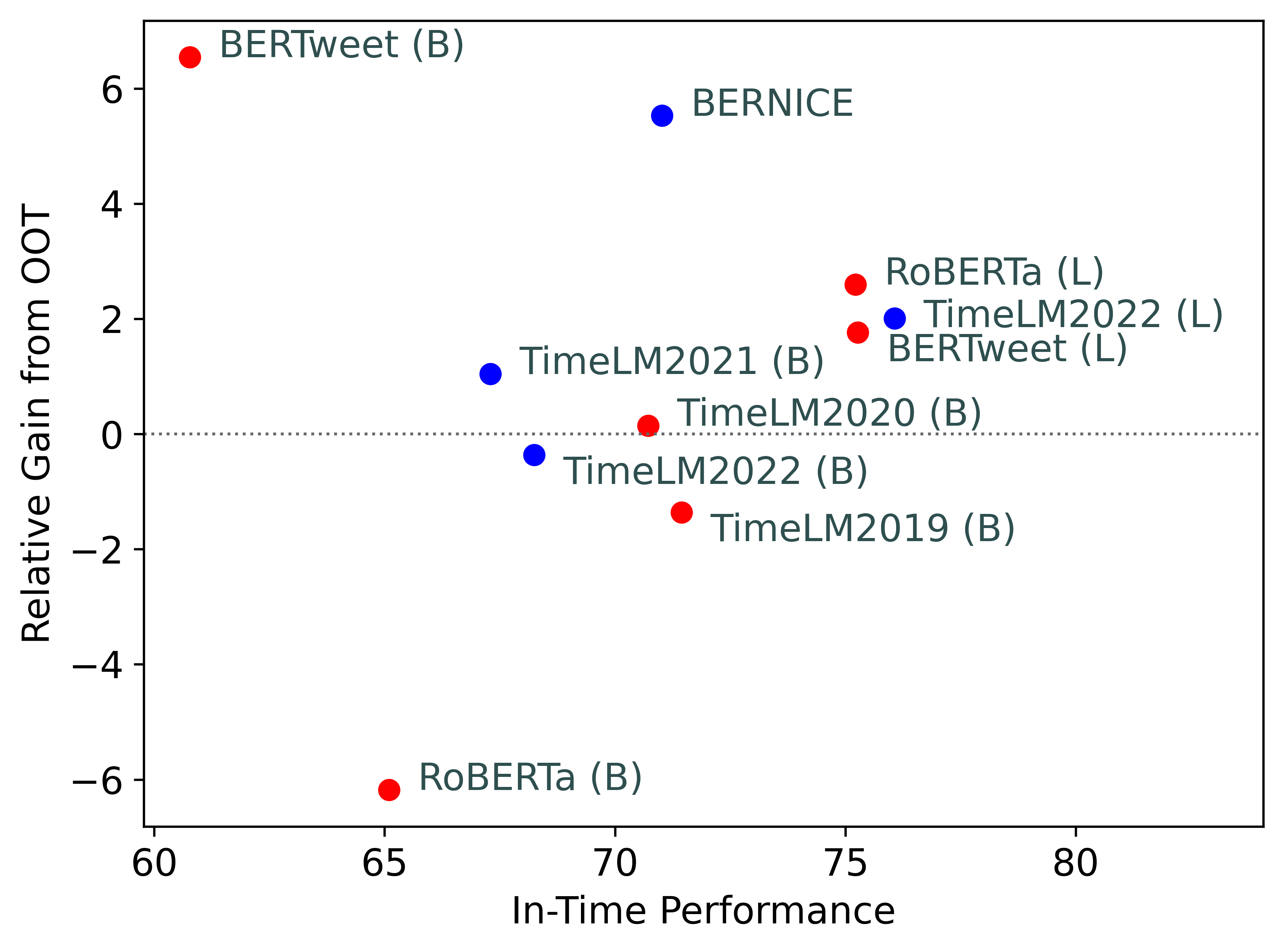}
         \caption{Topic classification.}
     \end{subfigure}
     \hfill
     \begin{subfigure}[b]{\columnwidth}
         \centering
         \includegraphics[width=\textwidth]{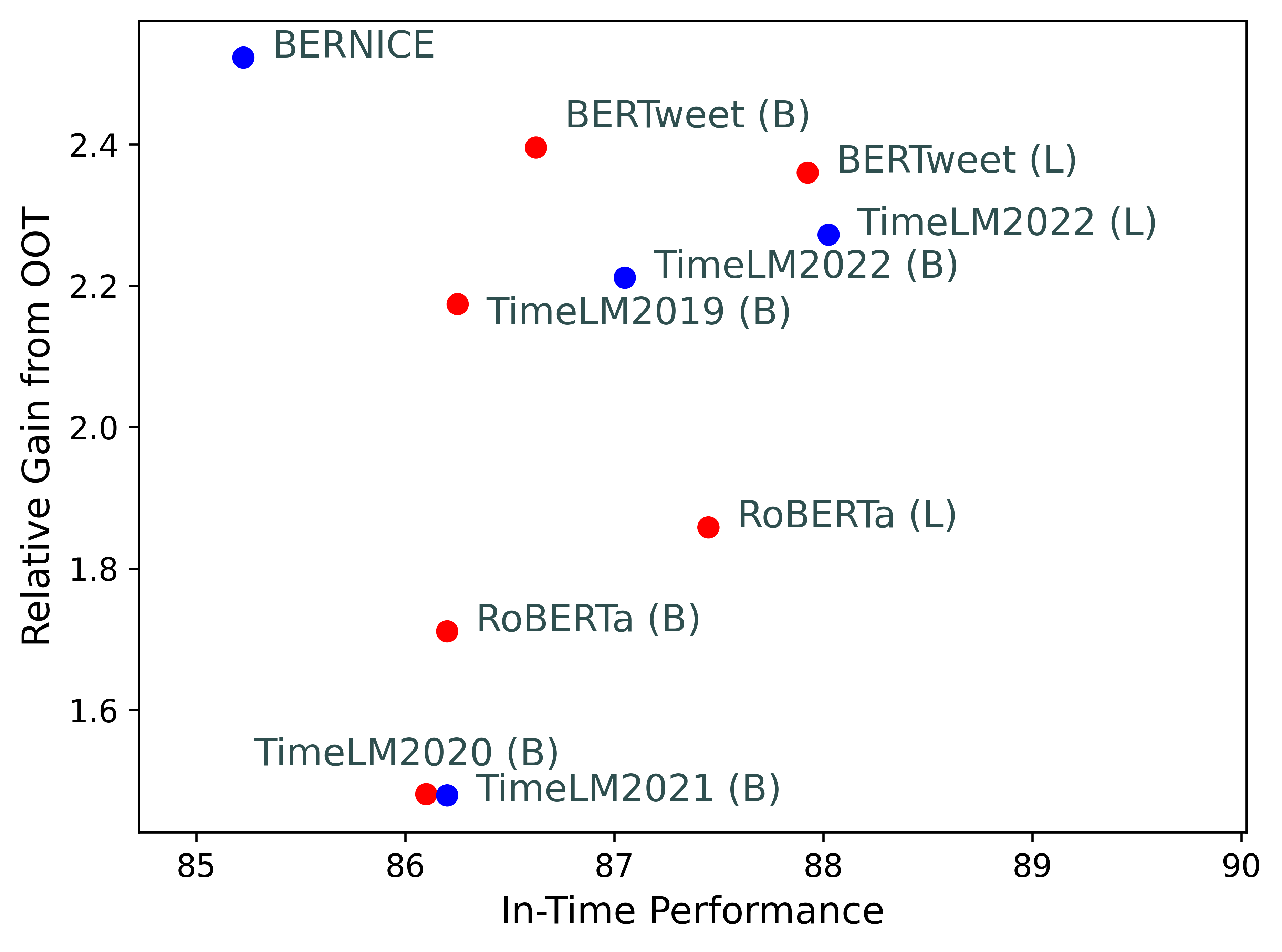}
         \caption{NED.}
     \end{subfigure}
     \hfill
     \begin{subfigure}[b]{\columnwidth}
         \centering
         \includegraphics[width=\textwidth]{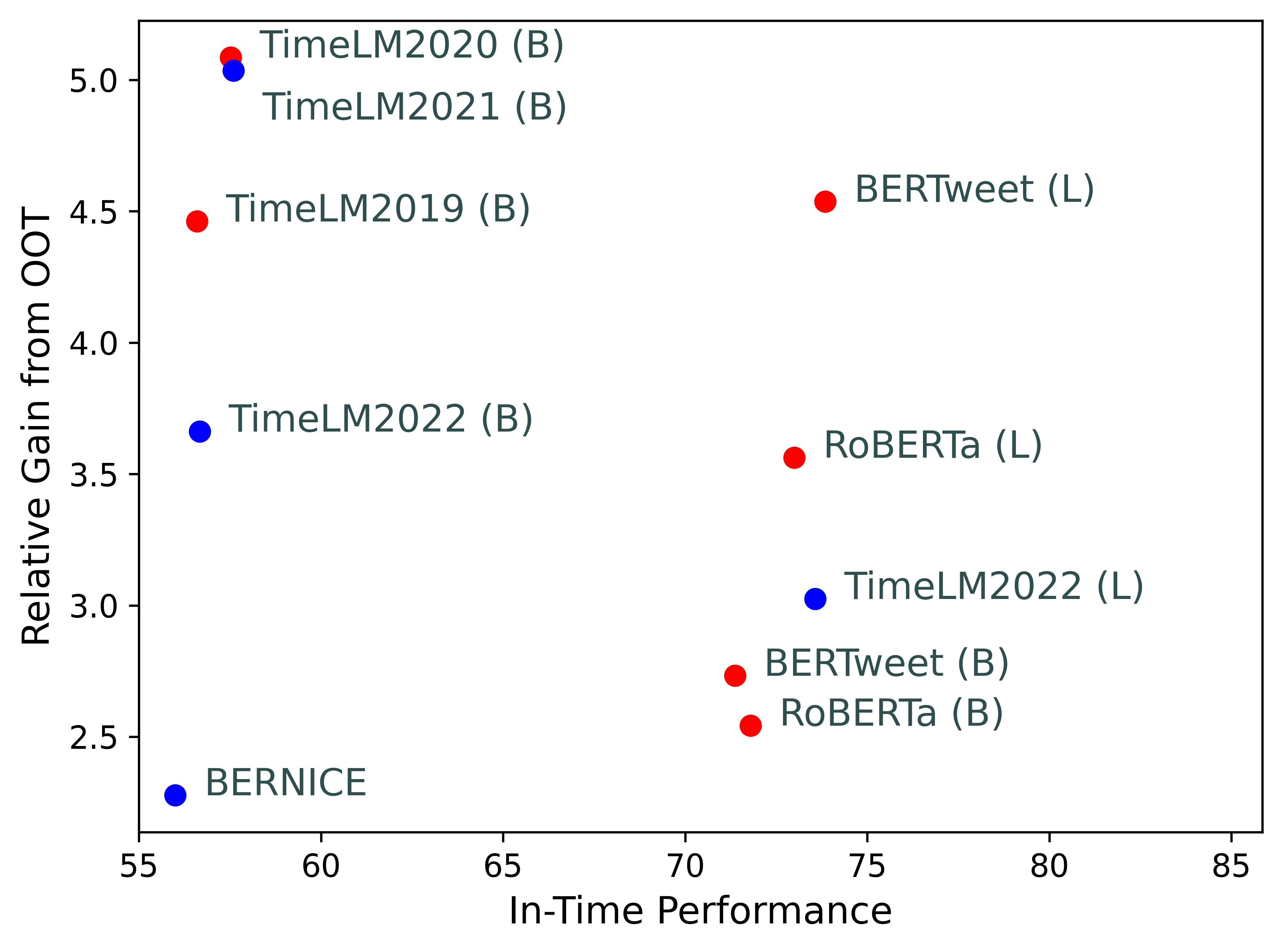}
         \caption{NER.}
     \end{subfigure}
\caption{Relative improvement (\%) from OOT to IT for each task (topic, NED and NER). LMs with pre-training corpus including the test period are in blue, and those without temporal overlap in red.}
\label{fig:result_relative}
\end{figure}

A possible direction to mitigate the temporal shift is to pre-train the LMs on the text from the test period, which does not require any labeling. Figure~\ref{fig:result_relative} visualizes the performance and relative IT improvement of LMs with/without pre-training corpus covering the test period of each task for topic classification/NED/NER\footnote{The test periods of hate speech detection and sentiment classification are covered by all the LMs we considered in the experiment.}. At a glance, we cannot observe see any relationship between the pre-training corpus and the performance. 
The averaged relative gains of the metrics from OOT within the LMs pre-trained on the test period and the others are 2.0 and 0.6 (topic classification), 3.5 and 3.8 (NER), and 2.1 and 1.9 (NED) respectively. Therefore, all models are affected by the temporal shift irrespective of the pre-training corpus date. This implies that the temporal shift cannot be robustly resolved by only adding data from the test period to the pre-training corpus, a conclusion that was also reached by \citeauthor{luu-etal-2022-time-custom} (\citeyear{luu-etal-2022-time-custom}).

\subsection{Effect of Label Distribution}

\begin{table}[t]
    \centering
    \scalebox{0.95}{
    \begin{tabular}{lrr}
    \toprule
    {} &  Original &  Balanced \\
    \midrule
    RoBERTa (B)    &       7.25 &     5.19 \\
    RoBERTa (L)    &       5.96 &    -0.95 \\
    BERTweet (B)   &       5.91 &     4.84 \\
    BERTweet (L)   &       2.88 &    -0.30 \\
    BERNICE        &       5.04 &     4.72 \\
    TimeLM2019 (B) &       4.80 &     4.16 \\
    TimeLM2020 (B) &       5.71 &     5.39 \\
    TimeLM2021 (B) &       4.51 &     5.52 \\
    TimeLM2022 (B) &       4.97 &     0.15 \\
    TimeLM2022 (L) &       4.94 &     1.89 \\ \midrule
    Average        &       5.20 &     3.06 \\
    \bottomrule
    \end{tabular}
        }
\caption{Comparisons of relative accuracy gain from OOT to IT between original (unbalanced) and balanced label distributions for hate speech detection.}
\label{tab:balance-experiment}
\end{table}

In supervised machine learning label distribution, the distribution of the binary label over the test instances, shifts can affect a model's performance. In this section, we analyse this potential effect when it comes to temporal shifts. For this, we rely on hate speech detection, which presents the largest decrease in performance from IT to OOT, with a different label distribution between training and test (see Figure~\ref{fig:label-dist-binary}). For the other tasks, the label distribution appears to be largely similar. To separate the effect of label distributional shift between IT and OOT from the temporal shift, we conduct a controlled experiment by balancing the label distribution of each IT training split to be the same as OOT training split. This is achieved by undersampling the size of the training set. Table~\ref{tab:balance-experiment} shows the results, where the average relative gain is still positive, although it becomes less dominant in balanced experiment. This highlights how label distribution may change over time and this itself have an effect in model performance. A similar finding was already discussed by \citeauthor{luu-etal-2022-time-custom} (\citeyear{luu-etal-2022-time-custom}).

\subsection{Qualitative Analysis}
In this analysis, we have a closer look on the test instances that are incorrect in OOT, turning to be correct in IT. To be precise, we sort the test instance in a single task based on the number of models where the error in OOT setting has been corrected in IT setting over all the random seeds. In other words, given a test instance, we check whether a model prediction is incorrect in OOT, but correct in the IT setting. This particular instance is counted as a correction. In total, we have 10 models with 3 independent runs with different random seed to construct the training data, so 30 would be the maximum number of corrections. For sentiment classification, hate speech detection, topic classification and NED, we simply count instance-level corrections. Given the complex nature of NER evaluation, we decided to only focus on the entity type predictions for this analysis.

\begin{table}[t]
\centering
\begin{tabular}{llr}
\toprule
Task      & Top corrected & Avg Top 10 \\
\midrule
NED       & 30/30 (100\%)                 & 30.0       \\
Hate      & 30/30 (100\%)                 & 30.0       \\
NER       & 28/30 (93.3\%)                & 24.0      \\
Sentiment & 19/30 (63.3\%)                & 13.6       \\
Topic     & 16/30 (53.3\%)                & 12.5      \\ \bottomrule
\end{tabular}
\caption{Top instances in terms of number of predictions corrected with an IT split. The second column indicates the top 10 average.}
\label{tab:tmp2}
\end{table}

\begin{table}[t]
\centering
\scalebox{0.8}{
\begin{tabular}{lp{90pt}rr}
\toprule
Task                  & Instance & Gold & Times corrected \\
\midrule
NED  & so cute how $<$Aoki$>$ describes Ida. "thinks about things seriously" \textit{(Japanese manga series)} & False                    & 30/30 (100\%)    \\ \cmidrule{2-4}
 & Will Ram \& $<$Priya$>$ go on a honeymoon it'll be a nice break for them (...) \#BadeAchheLagteHain2 \textit{(Indian actress)} & False                    & 30/30 (100\%)    \\ \midrule
Hate & \#MKR God Kat you are awful awful person. Oh you are humiliated? GOOD.                                                                                                       & False                    & 30/30 (100\%)    \\ \cmidrule{2-4}
 & \#katandandre gaaaaah I just want to slap her back to WA \#MKR & False & 30/30 (100\%) \\ \bottomrule
\end{tabular}
}
\caption{Two examples from the NED and hate speech detection datasets in which the prediction was corrected 100\% of the times with an IT split. For NED, the definition is provided in parenthesis and target word indicated between $<$ and $>$.}
\label{tab:tmp1}
\end{table}

Table \ref{tab:tmp2} shows the top instances in terms of IT corrections for each of the task. We can observed the marked differences across tasks, with NED and hate speech detection including instances which were corrected 100\% in the OOT setting. In fact, there are respectively 44 and 15 instances for which this is the case in these two tasks. Similarly for NER, the number of corrections is high. This is correlated with the main results of the paper (see Section \ref{sec:result}) which showed clear improvements for these tasks in the IT setting, but not for sentiment and and topic classification.

Finally, Table~\ref{tab:tmp1} shows some of these instances for NED and hate speech detection. In the case of NED, the tweets relate to two new TV series that were on air at test time (Japanese \textit{Kieta Hatsukoi} in the first example and Indian B\textit{ade Achhe Lagte Hain} in the second, both from 2021). This is similar to the hate speech detection in which the examples belong to the \textit{My Kitchen Rules TV} show. This highlights the event-driven nature of social media, and the importance of acquiring the background context for the specific task.

\section{Conclusion}
We proposed an evaluation method to assess the adaptability of LMs for temporal shifts on social media with five diverse downstream tasks including sentiment classification, NER, NED, hate speech detection, and topic classification. We have tested diverse LMs trained on Twitter under different temporal settings. The experimental results indicate that the adaptability gets consistently worse on entity or event-driven tasks (NED, NER, and hate speech detection) while the effect is limited in the other tasks. This conclusion was similar to previous work in more general domains, which observed a variation across different types of task when it comes to temporal degradation \cite{luu-etal-2022-time-custom,agarwal-nenkova-2022-temporal}. Finally, our analysis shows that pre-training on a corpus from the test period is not enough to solve the temporal shift issue, with performance still being degraded in comparison to models fine-tuned on the labeled dataset from the test period.

\section*{Limitations.}
Regardless of some similarities between Twitter and other streaming data such as news and other social media platforms being real-time and trend-driven, they can have different characteristics, and the results of our study may apply to Twitter exclusively. For our evaluation we rely on a single dataset for each of the tasks. Of course, these datasets are not a faithfully representation of the task and may contain their own biases. Therefore, even for the same task, the findings in this paper may differ if using a different dataset.

\section*{Ethical Statement.}
The datasets we used in the experiments are all from Twitter. Data has been anonymized (only information about legacy-verified users is kept) so that they do not contain any personal identifiable information (PII). We do not gather information from individual accounts but rely on aggregated information and metrics only. Please note that the text may contain sensitive content due to the nature of social media and the task, in particular hate speech detection.

\bibliography{anthology,anthology_p2,custom}



\end{spacing}

\end{document}